\documentclass[letterpaper, 10 pt, conference]{ieeeconf}  

\IEEEoverridecommandlockouts                              

\usepackage{cite}
\usepackage{amsmath,amssymb,amsfonts}
\usepackage{graphicx}
\usepackage{textcomp}
\usepackage{xcolor}
\usepackage{xspace}
\usepackage{booktabs}
\usepackage{multirow}
\let\labelindent\relax
\usepackage{enumitem}
\usepackage[breaklinks,colorlinks]{hyperref}
\usepackage{url}
\usepackage{physics}
\usepackage{colortbl} 
\usepackage{tikz}
\usepackage{bm}
\usepackage{subcaption}
\usepackage{amsmath}
\usepackage{svg}

\usepackage{pifont}
\newcommand{\cmark}{\ding{51}} 
\newcommand{\xmark}{\ding{55}} 
\newcommand{\method}{BRUM\xspace}
\newcommand{\dataset}{BRUM-dataset\xspace}
\newcommand{\tit}[1]{\smallskip\noindent\textbf{#1.}}
\newcommand{\ie}{\textit{i.e.}}

\title{\LARGE \bf
BRUM: Robust 3D Vehicle Reconstruction \\ from 360° Sparse Images 
}

\author{Davide Di Nucci$^{1}$, Matteo Tomei$^{2}$, Guido Borghi$^{1}$, Luca Ciuffreda$^{2}$, Roberto Vezzani$^{1}$, Rita Cucchiara$^{1}$ \\
\textsuperscript{1}University of Modena and Reggio Emilia, \textsuperscript{2}Prometeia
}

\def\BibTeX{{\rm B\kern-.05em{\sc i\kern-.025em b}\kern-.08em
    T\kern-.1667em\lower.7ex\hbox{E}\kern-.125emX}}
\begin{document}
\maketitle

\begin{abstract}
Accurate 3D reconstruction of vehicles is vital for applications such as vehicle inspection, predictive maintenance, and urban planning. Existing methods like Neural Radiance Fields 
and Gaussian Splatting 
have shown impressive results but remain limited by their reliance on dense input views, which hinders real-world applicability. This paper addresses the challenge of reconstructing vehicles from sparse-view inputs, leveraging depth maps and a robust pose estimation architecture to synthesize novel views and augment training data. Specifically, we enhance Gaussian Splatting by integrating a selective photometric loss, applied only to high-confidence pixels, and replacing standard Structure-from-Motion pipelines with the DUSt3R architecture to improve camera pose estimation. Furthermore, we present a novel dataset featuring both synthetic and real-world public transportation vehicles, enabling extensive evaluation of our approach. Experimental results demonstrate state-of-the-art performance across multiple benchmarks, showcasing the method’s ability to achieve high-quality reconstructions even under constrained input conditions. Code and data are publicly available at \url{https://aimagelab.ing.unimore.it/go/brum}.
\end{abstract}

\section{Introduction}

Novel view synthesis and 3D scene representation have been driven by techniques such as Neural Radiance Fields (NeRF)~\cite{mildenhall2021nerf} and Gaussian splatting (GS)~\cite{kerbl20233d} in the last few years. Both are capable of generating captures of complex scenes from input views with their corresponding camera poses: NeRF leverages an implicit representation through a neural network 
which models color and density of the scene as a function of  the spatial coordinates and viewing directions;
GS employs an explicit representation of the scene using a set of 3D Gaussian primitives with associated attributes (\textit{e.g.}, color, opacity).

While the focus of these methods has often been on general (both synthetic and real-world) scenes or human-centered tasks, their application to specific object classes -- \textit{e.g.}, the vehicle domain -- is still under-investigated. Accurate 3D vehicle reconstruction is essential for several critical applications. One primary motivation is automated \textit{vehicle inspection}~\cite{di2023carpatch,di2024kronc}, where precise 3D models enable detailed analysis of structural integrity, surface condition, and potential defects. In addition, 3D reconstruction facilitates tracking the status of vehicles over time, providing a digital record of wear, modifications, or damage, and supporting predictive maintenance. Autonomous driving is increasingly benefiting from novel view synthesis techniques, too~\cite{tonderski2024neurad,zhou2024drivinggaussian}.

Beyond individual monitoring, these models can also enhance fleet management systems by providing actionable insights into vehicle health and optimizing operations. Building on these advantages, the public transportation sector stands to benefit significantly from accurate reconstruction. Furthermore, in the context of smart cities, 3D models of public transportation vehicles can contribute to broader initiatives such as traffic management, urban planning, and environmental monitoring~\cite{turki2022mega,xie2024citydreamer,liu2025citygaussian}.

In practical scenarios, such as tracking a vehicle's condition over time (\textit{e.g.}, daily inspections for buses), capturing images quickly and consistently is crucial. While video recordings enable rapid data collection, they are often constrained by significant storage and processing requirements. Human-operated data collection could be significantly faster and more reliable if it required only a predetermined set of sparse images. Fixed camera setups offer a cost-effective alternative for image acquisition but inherently limit the number of available views to the number of cameras deployed.

In this work, we tackle the challenge of 3D vehicle reconstruction under the constraint of limited sparse views. Specifically, starting with a handful of scene captures and their associated camera parameters, we leverage ground truth or estimated depth maps to project view-dependent point clouds into synthetic camera poses. These new poses are generated by systematically rotating and translating the original camera positions within a constrained range, effectively augmenting the available views for downstream Gaussian splatting training. Importantly, for these views, the photometric loss is applied exclusively to pixels with high-confidence re-projection.

Camera position estimation for real-world scenes is traditionally performed using standard structure-from-motion (S\textit{f}M) pipelines, such as COLMAP~\cite{schonberger2016structure}.
Moreover, these methods often struggle or fail to converge when provided with only a few images or when image overlap is minimal~\cite{di2024kronc,truong2023sparf}. To address this limitation, we leverage the recently proposed DUSt3R architecture~\cite{wang2024dust3r} for estimating both camera poses and an initial point cloud.


We show that our proposed method achieves results comparable or even better than the state of the art in forward-facing 360° vehicle scenes from the Carpatch~\cite{di2023carpatch} and KRONC~\cite{di2024kronc} datasets, using as few as 4-8 images, without adding considerable computation time. To assess the effectiveness of our approach in large public transportation vehicle scenes, we also introduce the \dataset with both synthetic and realistic bus instances, highlighting promising results on it.

To sum up, our key contributions can be outlined as follows:
\begin{itemize}
     \item We enhance Gaussian splatting robustness in sparse-view forward-facing 360° vehicle scenes. Our method (\method) synthesizes novel images from coarse point clouds, effectively augmenting the input data. 
     \item We adopt DUSt3R as a replacement for the standard COLMAP and incorporate a masking strategy to exclude low-confidence pixels from the loss computation.
     \item We present the \dataset featuring 6 synthetic and 6 real public transportation vehicles. Our method achieves state-of-the-art performance on this dataset and other standard car inspection benchmarks.
\end{itemize}
\section{Related Work}

\tit{Novel View Synthesis}
Novel view synthesis aims to generate photorealistic renderings of a 3D scene from unseen viewpoints. In recent years, this challenge has been predominantly addressed by NeRF-like approaches~\cite{mildenhall2021nerf}, with various works proposing solutions to overcome the limitations of the original formulation. For faster training, \cite{muller2022instant} proposed to encode the 3D space using a multiresolution hash table, followed by a small MLP. Plenoxels~\cite{fridovich2022plenoxels} replaces neural networks with a voxel grid that directly stores density and spherical harmonic coefficients at each voxel, optimized using the standard MSE reconstruction loss. 
Other works focus on removing the dependency on S\textit{f}M preprocessing: \cite{lin2021barf,chen2023local} use bundle adjustment to refine camera parameters starting from noisy ones, \cite{bian2023nope} optimizes predicted depth maps distortion parameters and camera poses together with NeRF, \cite{chen2024neural} introduces a feature synthesizer to render a dense feature map based on a coarse absolute camera pose predicted by an absolute pose regressor, minimizing the error between this feature map and the output of a pre-trained feature extractor.
Recently, Gaussian splatting~\cite{kerbl20233d} has gained traction as a promising alternative to NeRF, providing advantages in training and rendering speed without compromising image quality. To address its shortcomings, the research community has undertaken various efforts. For instance, several studies have focused on reducing the storage footprint and memory requirements for deployment on edge devices, by compressing the Gaussian representation~\cite{morgenstern2023compact,niedermayr2024compressed,fan2023lightgaussian,zhang2025gaussianimage}. Other research has explored scene segmentation in 3D to enable editing capabilities~\cite{cen2023segment,hu2024semantic,chen2024gaussianeditor}. Moreover, text-to-3D scene generation is rapidly emerging as a prominent task, combining 3D Gaussian splatting with the generative power of 2D diffusion models~\cite{poole2022dreamfusion,zhou2025dreamscene360,chen2024text,liang2024luciddreamer}.

These methods highlight the capabilities of novel view synthesis techniques in efficiently reconstructing 3D scenes. However, they typically require capturing tens or hundreds of images to achieve accurate reconstructions, which can be challenging and costly. Our work focuses on addressing this limitation by targeting limited input images settings.

\tit{Few-shot Novel View Synthesis}
In literature, many studies have attempted to enhance the robustness of NeRF and Gaussian splatting under sparse-view settings. For instance, \cite{yu2021pixelnerf} introduces a method that leverages a fully-convolutional, pre-trained image encoder to predict a feature grid from the input image. The grid is then used to compute color and density at given spatial locations using NeRF. Similarly, IBRNet~\cite{wang2021ibrnet} proposes a ray-based transformer to aggregate information from neighboring views, enabling accurate predictions of color and density for novel target views. 
Other approaches focus on leveraging multi-view correspondences and geometric constraints among training views to improve reconstruction under sparse input conditions~\cite{truong2023sparf,lao2023corresnerf}. Depth supervision, derived from S\textit{f}M point clouds, monocular depth estimation, or depth completion models, has also been explored to enhance reconstruction quality~\cite{deng2022depth,roessle2022dense}.
The sparse-view problem similarly affects Gaussian splatting, prompting the development of several solutions. SplatFields~\cite{mihajlovic2025splatfields} identifies overfitting to training views due to low spatial autocorrelation between splats and introduces a neural framework to promote feature sharing among nearby primitives. CoherentGS~\cite{paliwal2025coherentgs} employs monocular depth to initialize 3D Gaussians and refines them using single- and multi-view regularization based on depth and optical flow constraints. 
Others use robust pre-trained image matchers to incorporate multi-view matching priors~\cite{yin2024fewviewgs,peng2024structure}.

Unlike existing methods that often rely on sophisticated architectures with considerable training or inference time overhead, \method involves minimal additional complexity. Our work focuses on 360° forward-facing vehicle scenes with sparse views, a scenario typically explored in synthetic settings but rarely addressed in real-world environments.

\begin{figure*}[th!]
     \centering
     \begin{subfigure}[b]{0.45\linewidth}
         \centering
         \includegraphics[width=\textwidth]{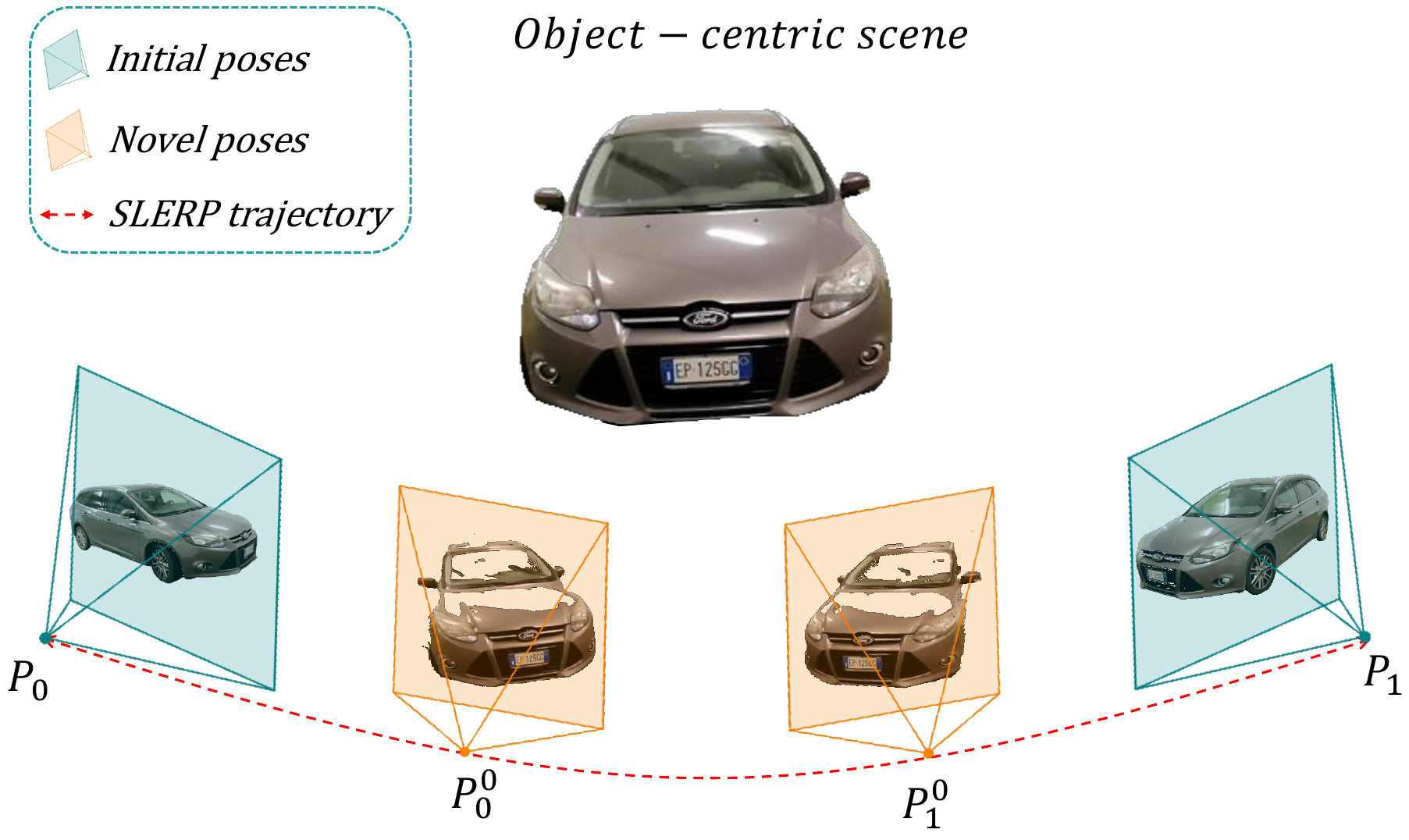}
         \caption{}
         \label{fig:method_1}
     \end{subfigure}
     \medspace
     \begin{subfigure}[b]{0.45\linewidth}
         \centering
         \includegraphics[width=\textwidth]{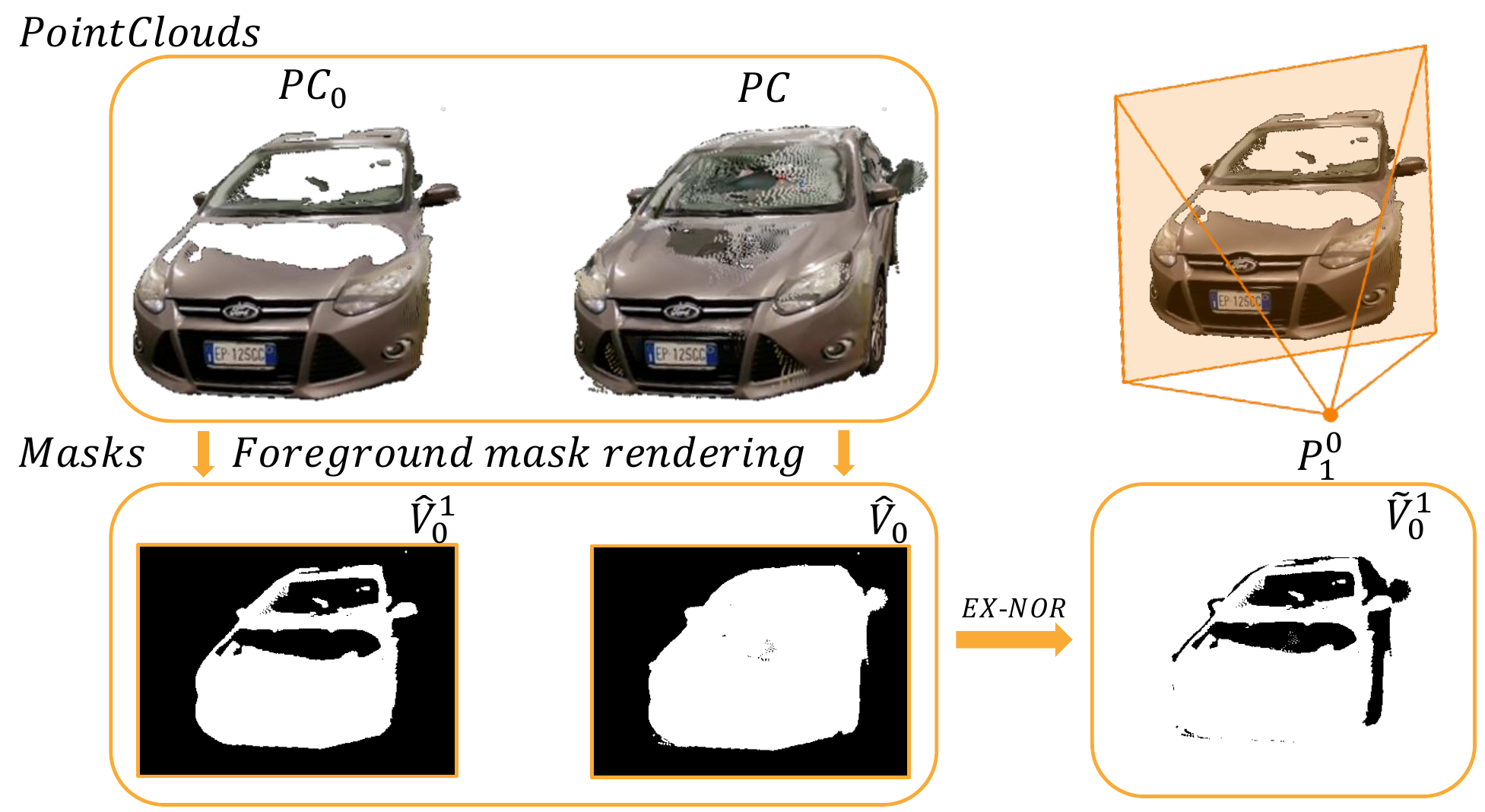}
         \caption{}
         \label{fig:method_2}
     \end{subfigure}
        \caption{ (a) illustrates an object-centric scene, where ground-truth camera poses are augmented along a SLERP (Sec.~\ref{sec:augmentation}) trajectory to generate novel poses. (b) shows how to obtain the final foreground mask $\tilde{V}_0^1$, starting from point clouds and intermediate masks $\hat{V}_0^1$ and $\hat{V}_0$.}
        \label{fig:method}
\end{figure*}

\section{Method}


\subsection{Augmenting the available training views}
\label{sec:augmentation}
The input to our algorithm is a set of $N$ sparse images, denoted as $\mathcal{I} = \{I_i\}_{i=1}^N$, capturing an object-centric scene. We assume depth maps for each input image $\mathcal{D} = \{D_i\}_{i=1}^N$ being available, together with corresponding extrinsic camera parameters $\mathcal{P} = \{P_i\}_{i=1}^N$, with $P_i=[R_i, \mathbf{t}_i]$, where $R_i \in SO(3)$ and $\mathbf{t}_i \in \mathbb{R}^3$ are the rotation matrix and translation vector of camera $i$, respectively, relative to a common world coordinate system. We also suppose known camera intrinsic parameters, shared among all views.

\tit{Sampling novel poses}
For the $i^{th}$ camera, we propose to generate $M$ different novel poses starting from its parameters, $P_i = [R_i, \mathbf{t}_i]$, by applying rotations and translations within a limited range. To facilitate this, we first identify the closest camera, indexed as $k$, among the other $N-1$ cameras, characterized by its parameters $P_k = [R_k, \mathbf{t}_k]$. The pair of cameras ${P_i, P_k}$ defines the starting and ending points for interpolation. 
To smoothly transition between these two poses, we adopt the Spherical Linear Interpolation (SLERP)~\cite{shoemake1985animating} algorithm. SLERP interpolates between two unit quaternions along the shortest geodesic path on the unit sphere in $\mathbb{R}^4$, ensuring constant-speed rotation transitions. This approach maintains the object-centric nature of the scene while enabling precise and smooth interpolation for generating novel views.

$P_i$ and $P_k$ are first both converted to quaternions $\mathbf{q}_i$ and $\mathbf{q}_k$, then SLERP computes the intermediate quaternion $\mathbf{q}_h$ for interpolation parameter $h \in [0, 1]$ as follows:
\begin{equation}
\label{eq:slerp}
\mathbf{q}_h = \frac{\sin((1 - h) \theta)}{\sin\theta} \mathbf{q}_i + \frac{\sin(h \theta)}{\sin\theta} \mathbf{q}_k,
\end{equation}
where $\theta$ is the angle subtended by the arc between $\mathbf{q}_i$ and \(\mathbf{q}_k\). Finally, $\mathbf{q}_h$ is converted back to $[R_h, \mathbf{t}_h]$. In practice, instead of relying solely on the nearest camera, we independently apply SLERP between the $i^{th}$ camera and its two closest cameras. This enables interpolation along two distinct arcs, enhancing the diversity of the generated poses. The number of generated poses can be controlled by sampling various values of $h$, while constraining $h$ within a specific range of values allows for limiting the deviation from the $i^{th}$ camera. This results in a total of $M$ new poses $\hat{\mathcal{P}}_i = \{\hat{P}_i^j\}_{j=1}^M$ sampled for camera $i$.

\tit{Generating novel views}
Starting from poses $\hat{\mathcal{P}}_i$, by leveraging the RGB information in $I_i$ and the depth $D_i$, we aim to produce synthetic novel views of the scene. Specifically, we first back-project pixels from $I_i$ to 3D using the depth map $D_i$, obtaining a 3D point cloud $PC_i$:
\begin{equation}
\label{eq:point_cloud_i}
PC_i = \pi^{-1}_{i}(I_i, D_i),
\end{equation}
where $\pi_i^{-1}$ represents the back-projection from the 2D image plane defined by $P_i$ to the common 3D world’s reference system. The point cloud $PC_i$ is then warped to the novel views defined by camera poses $\hat{\mathcal{P}}_i$. Given a synthetic camera pose $\hat{P}_i^j \in \hat{\mathcal{P}}_i$, we project $PC_i$ to its image plane as follows:
\begin{equation}
\hat{I}_i^j = \pi_{j}(PC_i),
\end{equation}
where $\pi_j$ is the projection from the common 3D world’s reference system to the 2D image plane defined by $\hat{P}_i^j$. In practice, while we use the na\"ive back-projection for $\pi^{-1}$ as a one-to-one mapping from 2D to 3D, we adopt the solution proposed by~\cite{wiles2020synsin} for projection $\pi$. Specifically, a 3D point $p$ from $PC_i$ is splatted onto a 2D disk with radius $r$ and center $p_c$, and its influence on a pixel $u$ in $\hat{I}_i^j$ is inversely proportional to the 2D Euclidean distance between $u$ and the disk's center $p_c$:
\begin{equation}
\label{eq:weights}
w(p, u) = 
\begin{cases}
    0 & \text{if } \| p_c - u \|_2 > r \\
    1 - \frac{\| p_c - u \|}{r^2} & \text{otherwise,}
\end{cases}
\end{equation}
where $w(p,u)$ represents a weight quantifying how much the 3D point $p$ affects pixel $u$.
As in~\cite{wiles2020synsin}, the projected points are then stored in a z-buffer, sorted by their distance from the new camera pose $\hat{P}_i^j$ and only the $K$ closest points are retained. Finally, alpha over-compositing is adopted for points accumulation. The weighting scheme helps ensure accurate novel views, particularly in cases of overlapping or occluded points, and guarantees smoother renderings.
 
  Binary foreground masks \( \hat{\mathcal{V}}=\{\hat{V}_i^j\}_{j=1}^M \) are also gathered, identifying the pixels that have been successfully projected from $PC_i$ to $\hat{I}_i^j$. 

\subsection{Training objective}
\label{sec:loss}
We can now exploit the original $N$ images $\mathcal{I} = \{I_i\}_{i=1}^N$ together with the novel $N \times M$ images $\hat{\mathcal{I}} = \{\{\hat{I}_i^j\}_{j=1}^M\}_{i=1}^N$ generated from the original ones, for downstream Gaussian splatting training.
We recall the original Gaussian splatting objective in the following equation:
\begin{equation}
\label{eq:gs_loss}
\mathcal{L} = (1-\lambda)\mathcal{L}_1(\tilde{I}, I) + \lambda\mathcal{L}_{SSIM}(\tilde{I}, I),
\end{equation}
which is a combination of the L1 and SSIM~\cite{wang2004image} losses between rendered ($\tilde{I}$) and ground truth images ($I$).

First, for the 3D Gaussians optimization, we exclude pixels from the generated images that are not accurately projected from the 3D point cloud, as indicated by the foreground masks. However, in object-centric scenes - where objects are segmented, and the background is removed - simply masking out these pixels in the loss can inadvertently exclude background Gaussians from optimization. This happens because background points are absent from the 3D point cloud and are therefore not reported in the foreground masks. As a result, background Gaussians would remain unoptimized, leading to poor 3D reconstruction where the background is only initialized but never refined. To address this, we distinguish between background pixels and those that are actually not reprojected. For each foreground mask in $\hat{\mathcal{ V}} = \{\{\hat{V}_i^j\}_{j=1}^M\}_{i=1}^N$, we compute its \textit{exclusive nor} with the foreground mask obtained by rendering the complete point cloud $PC=\bigcup_{i=1}^N PC_i$ from the same camera pose:
\begin{equation}
\label{eq:xnor}
\tilde{V}_i^j = 1 - (\hat{V}_i^j \oplus \hat{V}_i),
\end{equation}
where $\hat{V}_i$ is the foreground of $PC$ rendered from $\hat{P}_i^j$ and $\oplus$ is the XOR operator. We consider only pixels retained in $\tilde{V}_i^j$ in the final loss.
This approach retains background pixels while excluding those not reprojected from $PC_i$, ensuring a more accurate optimization. More details in Figure \ref{fig:method}.

Second, we associate a \textit{reliability} measure with each pixel in the generated images. Beyond masking-out pixels that are incorrectly projected from 3D, we apply weights to the remaining pixels using the influence values computed in Eq.~\ref{eq:weights}. Intuitively, a pixel largely affected by multiple 3D points is considered more reliable, and its weight in the loss function increases. Formally, for pixel $u$ in $\hat{I}_i^j$, its loss weight becomes:
\begin{equation}
\label{eq:weights_sum}
w_K(u) = \sum_{k=1}^K{w(p_k, u)},
\end{equation}
where $p_k$ is the $k^{th}$ point from point cloud $PC_i$ among the $K$ used for alpha over-compositing, as mentioned in Sec.~\ref{sec:augmentation}. This leads to weights $\hat{W}_i^j$ for image $\hat{I}_i^j$, after computing the above equation for all the pixels.
$\hat{W}_i^j$ is further normalized as:
\begin{equation}
\label{eq:weights_sum_norm}
\tilde{W}_i^j = \frac{\hat{W}_i^j - \min(\hat{W}_i^j)}{\max(\hat{W}_i^j) - \min(\hat{W}_i^j)}
\end{equation}

Finally, as the SSIM loss evaluates the structural similarity at the image level rather than on a per-pixel basis, and given that generated images are often less reliable and may lack consistent overall structure, we exclude the SSIM loss from the optimization process for generated images.

Our overall objective for generated images $\hat{\mathcal{I}}$ in Gaussian splatting training is:
\begin{equation}
\label{eq:final_loss}
    \mathcal{L} =
    \begin{cases}
    (1-\lambda)\mathcal{L}_1(\tilde{I}_i, I_i) + \lambda\mathcal{L}_{SSIM}(\tilde{I}_i, I_i) &, \forall \, I_i \in \mathcal{I} \\[0.5em]
    \displaystyle\frac{\sum_{u} \tilde{V}_i^j(u) \cdot \tilde{W}_i^j(u) \cdot |\tilde{I}_i^j(u) - \hat{I}_i^j(u)|}{\sum_{u} \tilde{V}_i^j(u)} &, \forall \, \hat{I}_i^j \in \mathcal{\hat{\mathcal{I}}}, \\
\end{cases}
\end{equation}
where the standard Gaussian splatting loss is applied to the original images, while the weighted and masked L1 loss is used for the generated images. Here $\tilde{I}_i$ and $\tilde{I}_i^j$ denote Gaussian splatting renderings from $P_i$ and $P_i^j$, respectively.

\subsection{Preprocessing for real-world scenes}
\label{sec:DUSt3R}
For real-word scenes, obtaining camera poses $\mathcal{P}$ and depth maps $\mathcal{D}$ is often challenging due to the limited accessibility of precise sensors and the expense associated with specialized imaging devices. Standard Gaussian splatting and NeRF pipelines usually rely on structure-from-motion as a pre-processing step to estimate both camera parameters and an initial point cloud. However, commonly used S\textit{f}M methods, such as COLMAP~\cite{schonberger2016structure}, often struggle to converge in sparse-view scenarios with a limited number of images.

To address this challenge, we adpot DUSt3R~\cite{wang2024dust3r} in place of COLMAP. DUSt3R processes image pairs using a shared ViT~\cite{dosovitskiy2020image} encoder to extract representations, which are then fed into two Transformer-based decoders~\cite{vaswani2017attention} equipped with cross-attention layers. These decoders predict two aligned point clouds with associated per-point confidence through regression heads. A subsequent global optimization step allows DUSt3R to determine, for each image $I_i$, the corresponding camera pose $P_i$, point cloud $PC_i$, and scale-aligned depth map $D_i$ (derived from the \textit{z}-coordinate of the predicted point cloud), as well as a confidence map $C_i$.

Since our focus is vehicle reconstruction, we further refine the images by leveraging Segment Anything~\cite{kirillov2023segment} to remove backgrounds. This step generates a set of segmentation masks, \( \mathcal{F} = \{F_i\}_{i=1}^N \). 

The overall pipeline is kept consistent with Sec.~\ref{sec:augmentation} and~\ref{sec:loss}, except for two modifications:
\begin{itemize}
    \item Each image $I_i$ is initially filtered using the segmentation mask, yielding $I_i=I_i \odot F_i$.
    \item The point cloud $PC_i$, obtained from Eq.~\ref{eq:point_cloud_i}, is refined based on both the segmentation mask and the confidence map, as follows:
    \begin{equation}
    \label{eq:point_cloud_filter}
        PC_i=\{p_k \in PC_i \mid F_i(u_k)=1 \text{ and } C_i(u_k)>c\},
    \end{equation}
    where $u_k$ is the 2D pixel location corresponding to the 3D point $p_k$ in $PC_i$, and $c$ is a fixed confidence threshold.
\end{itemize}

\definecolor{myrowcolor}{HTML}{F5F5F5}
\definecolor{first}{HTML}{dec546}
\definecolor{second}{HTML}{d7d7d7}
\definecolor{third}{HTML}{7e4205}
\begin{table*}[ht]
\centering
\caption{Quantitative results averaged over the \dataset and CarPatch scenes. }
\begin{tabular}{l|ccc|c||ccc|c}
\toprule
& \multicolumn{4}{c||}{\textbf{\dataset}} & \multicolumn{4}{c}{\textbf{CarPatch}} \\
\textbf{Method} & PSNR~$\uparrow$ & SSIM~$\uparrow$ & LPIPS~$\downarrow$ & AVGE~$\downarrow$ & PSNR~$\uparrow$ & SSIM~$\uparrow$ & LPIPS~$\downarrow$ & AVGE~$\downarrow$\\
\midrule
3DGS~\cite{kerbl20233d} &
21.08 \tikz\draw[second,fill=second,opacity=0.,fill opacity=0.](0,0)circle(.5ex); &
0.861 \tikz\draw[second,fill=second,opacity=0.,fill opacity=0.](0,0)circle(.5ex); &
0.132 \tikz\draw[second,fill=second,opacity=0.,fill opacity=0.](0,0)circle(.5ex); &
0.073 \tikz\draw[second,fill=second,opacity=0.,fill opacity=0.](0,0)circle(.5ex); & 
22.42 \tikz\draw[second,fill=second,opacity=0.,fill opacity=0.](0,0)circle(.5ex); &
0.867 \tikz\draw[second,fill=second,opacity=0.,fill opacity=0.](0,0)circle(.5ex); &
0.122 \tikz\draw[second,fill=second,opacity=0.,fill opacity=0.](0,0)circle(.5ex); & 
0.063 \tikz\draw[second,fill=second,opacity=0.,fill opacity=0.](0,0)circle(.5ex);  \\
\rowcolor{myrowcolor}
DNGaussians~\cite{li2024dngaussian} &
17.33 \tikz\draw[second,fill=second,opacity=0.,fill opacity=0.](0,0)circle(.5ex); &
0.617 \tikz\draw[second,fill=second,opacity=0.,fill opacity=0.](0,0)circle(.5ex); &
0.187 \tikz\draw[second,fill=second,opacity=0.,fill opacity=0.](0,0)circle(.5ex); &
0.129 \tikz\draw[second,fill=second,opacity=0.,fill opacity=0.](0,0)circle(.5ex); &
21.23 \tikz\draw[second,fill=second,opacity=0.,fill opacity=0.](0,0)circle(.5ex); &
0.824 \tikz\draw[second,fill=second,opacity=0.,fill opacity=0.](0,0)circle(.5ex); &
0.144 \tikz\draw[second,fill=second,opacity=0.,fill opacity=0.](0,0)circle(.5ex); & 
0.077 \tikz\draw[second,fill=second,opacity=0.,fill opacity=0.](0,0)circle(.5ex);  \\
SplatFields~\cite{mihajlovic2025splatfields} &
23.80 \tikz\draw[second,fill=second,opacity=0.,fill opacity=0.](0,0)circle(.5ex); &
0.887 \tikz\draw[second,fill=second,opacity=0.,fill opacity=0.](0,0)circle(.5ex); &
0.118 \tikz\draw[second,fill=second,opacity=0.,fill opacity=0.](0,0)circle(.5ex); &
0.055 \tikz\draw[second,fill=second,opacity=0.,fill opacity=0.](0,0)circle(.5ex); &
23.39 \tikz\draw[second,fill=second,opacity=0.,fill opacity=0.](0,0)circle(.5ex); &
0.889 \tikz\draw[second,fill=second,opacity=0.,fill opacity=0.](0,0)circle(.5ex); &
0.116 \tikz\draw[second,fill=second,opacity=0.,fill opacity=0.](0,0)circle(.5ex); & 
0.056 \tikz\draw[second,fill=second,opacity=0.,fill opacity=0.](0,0)circle(.5ex);  \\
\rowcolor{myrowcolor}
\method  &
\textbf{24.65} \tikz\draw[second,fill=second,opacity=0.,fill opacity=0.](0,0)circle(.5ex); &
\textbf{0.917} \tikz\draw[second,fill=second,opacity=0.,fill opacity=0.](0,0)circle(.5ex); &
\textbf{0.083} \tikz\draw[second,fill=second,opacity=0.,fill opacity=0.](0,0)circle(.5ex); &
\textbf{0.043} \tikz\draw[second,fill=second,opacity=0.,fill opacity=0.](0,0)circle(.5ex); &
\textbf{25.57} \tikz\draw[second,fill=second,opacity=0.,fill opacity=0.](0,0)circle(.5ex); &
\textbf{0.911} \tikz\draw[second,fill=second,opacity=0.,fill opacity=0.](0,0)circle(.5ex); &
\textbf{0.079} \tikz\draw[second,fill=second,opacity=0.,fill opacity=0.](0,0)circle(.5ex); & 
\textbf{0.040} \tikz\draw[second,fill=second,opacity=0.,fill opacity=0.](0,0)circle(.5ex);  \\
\bottomrule
\end{tabular}
\label{tab:tab_synt_results}
\end{table*}

\section{Experiments}  
All experiments were conducted on a TITAN RTX GPU.  

\tit{Competitor Implementation} We compare {\method}'s performance in both synthetic and real-world scenarios against Gaussian Splatting and two state-of-the-art architectures designed for sparse input settings: DNGaussians~\cite{li2024dngaussian} and SplatFields~\cite{mihajlovic2025splatfields}.
To ensure fair comparisons, we introduced minor yet essential modifications to these methods to better align them with the requirements of our use case. Specifically, for SplatFields, the original implementation proposes scaling the initial learning rate by a constant value. However, we found that dynamically adjusting the learning rate based on the characteristics of each training scene was necessary to achieve optimal performance in our experimental setup. In the case of DNGaussians,
instead of optimizing parameters individually for each scene, we employed the default settings provided for the LEGO scene in the Blender dataset. This approach ensured the reproducibility of the results over different scenarios.


\tit{Evaluation Metrics}
We evaluate our method using standard visual quality metrics: Peak Signal-to-Noise Ratio (PSNR), Structural Similarity Index Measure (SSIM)~\cite{wang2004image}, and Learned Perceptual Image Patch Similarity (LPIPS)~\cite{zhang2018unreasonable}, which assess reconstruction fidelity, structural consistency, and perceptual quality, respectively. To provide a more intuitive and comprehensive comparison, we also report the Average Error (AVGE)~\cite{niemeyer2022regnerf}. This metric integrates multiple aspects of visual quality by combining $MSE = 10^{-\text{PSNR}/10}$, $\sqrt{1 - \text{SSIM}}$, and LPIPS as a geometric mean, capturing both pixel-wise and perceptual errors in a unified score.

\tit{Rendering} For the rendering process, we use the \texttt{PointsRenderer} module from PyTorch3D. The number of points contributing to a pixel's color (\ie~the $K$ parameter introduced in Sec.~\ref{sec:augmentation}) is set to 16 for both synthetic and real-world datasets to ensure smooth and consistent outputs. The $r$ parameter (Eq. \ref{eq:weights}), which controls the projected size of points, is assigned a value of 0.003 for synthetic data and 0.1 for real-world data, allowing for an optimal balance between preserving fine details and minimizing overlapping effects. These parameter configurations facilitate high-quality renderings with minimal visual artifacts.

\subsection{The \dataset} \label{sec:dataset}
Given the lack of data in literature representing public transportation vehicles, we publicly release the \dataset. This dataset comprises 12 scenes including 6 real-world 360° captures of buses and 6 synthetic bus models. For both settings, the buses are labeled as bus\_1 through bus\_6. 

\begin{figure}[t]
    \centering
    \includegraphics[page=2, width=0.9\linewidth]{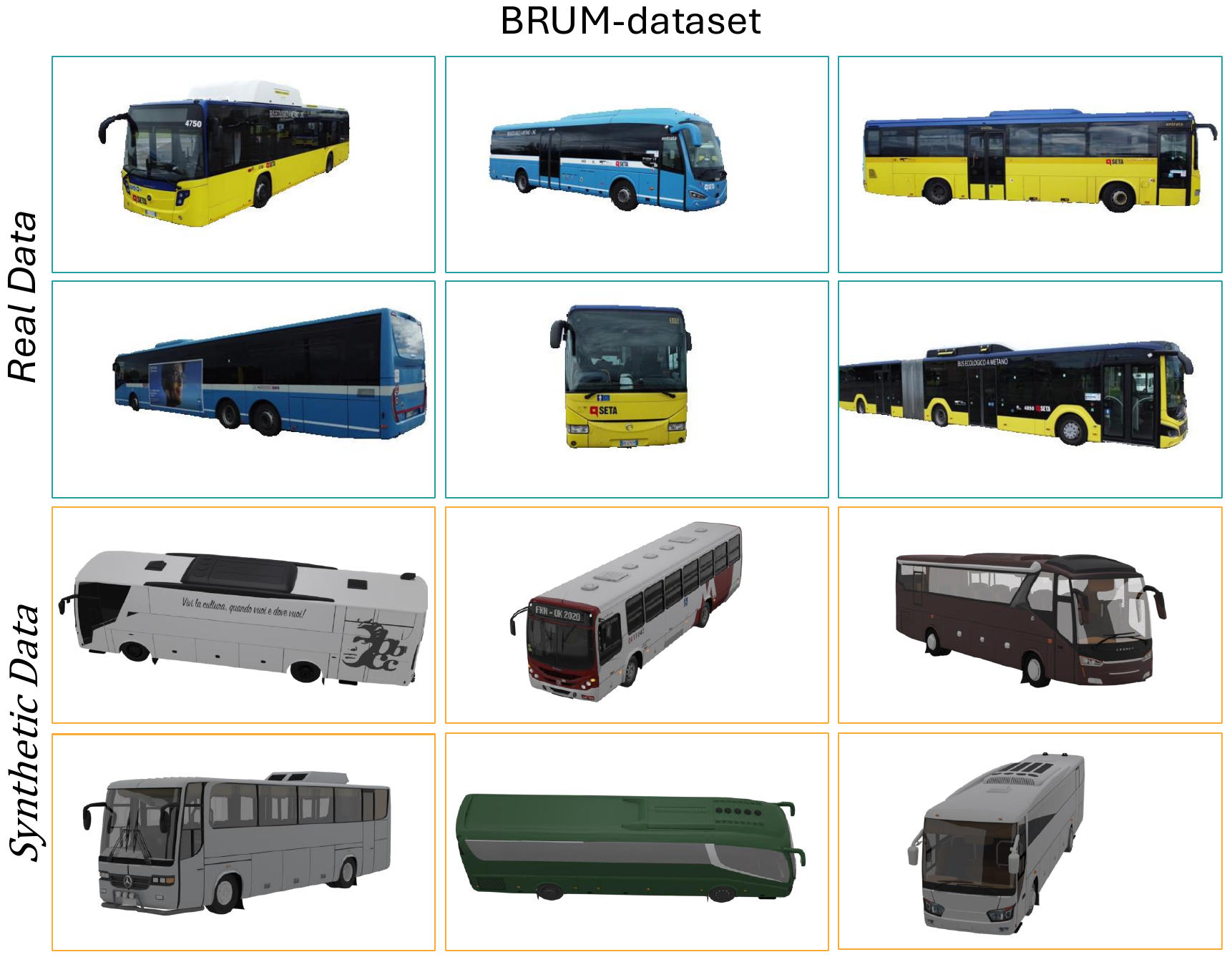}
    \caption{Overview of the \dataset.}
    \label{fig:dataset}
    \vspace{-.3cm}
\end{figure}

\tit{Synthetic} The \dataset consists of six distinct synthetic scenes, each featuring a unique 3D bus model.
All the 3D models have been downloaded from Sketchfab\footnote{\url{https://sketchfab.com}} and 3DExport\footnote{\url{https://3dexport.com}}.
These scenes were generated using Blender\footnote{\url{http://www.blender.org}}, inspired by the Google Blender dataset setup~\cite{mildenhall2021nerf}. This approach allowed precise control over lighting conditions and camera viewpoints, providing a proof of concept for evaluating our method under challenging conditions.
Each scene was designed with realistic rendering settings: the vehicle was positioned at the origin $(0,0,0)$, surrounded by nine lights with varying emission strengths to produce realistic shadows and reflections. Objects were resized to match real-world dimensions, the camera and lighting were configured to mimic real-world environments. Each scene includes 100 training images and 200 test images, accompanied by ground truth depth annotations and camera positions. For training, the camera was randomly positioned on a hemisphere above the ground, with rotation angles sampled uniformly. Instead, test images were captured with the camera at a fixed height, rotating around the Z-axis in steps of $\frac{2\pi}{\#test\_views}$ radians per frame.

\tit{Real-World} The \dataset also comes with six distinct real bus scenes, selected to represent a diverse range of designs and manufacturers. The data collection was performed in a controlled environment using a DJI MINI 2 SE drone. For each bus, a 360° video was captured by flying multiple laps around the vehicle at varying altitudes. The initial lap was performed at eye level, followed by subsequent laps with the altitude gradually increased by approximately 3 meters, ensuring thorough coverage of each bus's structure. The original videos were recorded at 30 fps with a resolution of $1920\times1080$. For compatibility with novel view synthesis techniques, individual frames were extracted and sub-sampled. For each bus, a masked version of all frames is provided as mentioned in \ref{sec:DUSt3R}.


\begin{figure}[t]
    \centering
    \includegraphics[width=\linewidth]{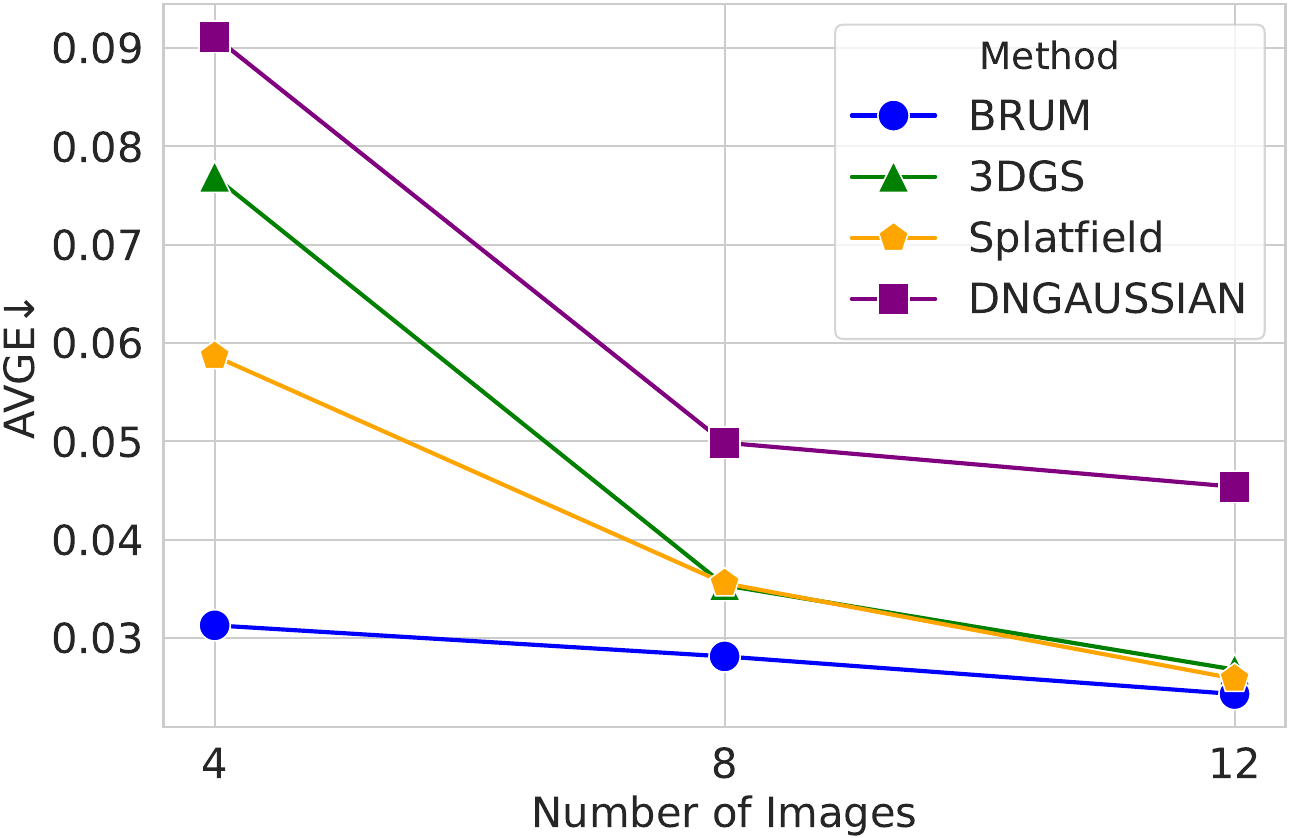}
    \caption{AVGE results by varying number of training images.}
    \label{fig:n_images_plot}
    \vspace{-.3cm}
\end{figure}

\definecolor{myrowcolor}{HTML}{F5F5F5}
\definecolor{first}{HTML}{dec546}
\definecolor{second}{HTML}{d7d7d7}
\definecolor{third}{HTML}{7e4205}
\begin{table*}[t]
\centering\caption{Quantitative results averaged over the KRONC and \dataset scenes.}
\begin{tabular}{l|ccc|c||ccc|c}
\toprule
& \multicolumn{4}{c||}{\textbf{KRONC}} & \multicolumn{4}{c}{\textbf{\dataset}} \\
\textbf{Method} & PSNR~$\uparrow$ & SSIM~$\uparrow$ & LPIPS~$\downarrow$ & AVGE~$\downarrow$ & PSNR~$\uparrow$ & SSIM~$\uparrow$ & LPIPS~$\downarrow$ & AVGE~$\downarrow$ \\
\midrule
3DGS~\cite{kerbl20233d} &
19.44 \tikz\draw[second,fill=second,opacity=0.,fill opacity=0.](0,0)circle(.5ex); &
0.764 \tikz\draw[second,fill=second,opacity=0.,fill opacity=0.](0,0)circle(.5ex); &
0.153 \tikz\draw[second,fill=second,opacity=0.,fill opacity=0.](0,0)circle(.5ex); &
0.094 \tikz\draw[second,fill=second,opacity=0.,fill opacity=0.](0,0)circle(.5ex); &
19.65 \tikz\draw[second,fill=second,opacity=0.,fill opacity=0.](0,0)circle(.5ex); &
0.734 \tikz\draw[second,fill=second,opacity=0.,fill opacity=0.](0,0)circle(.5ex); &
0.192 \tikz\draw[second,fill=second,opacity=0.,fill opacity=0.](0,0)circle(.5ex); &
0.102 \tikz\draw[second,fill=second,opacity=0.,fill opacity=0.](0,0)circle(.5ex); \\
\rowcolor{myrowcolor}
DNGaussians~\cite{li2024dngaussian} &
17.09 \tikz\draw[second,fill=second,opacity=0.,fill opacity=0.](0,0)circle(.5ex); &
0.704 \tikz\draw[second,fill=second,opacity=0.,fill opacity=0.](0,0)circle(.5ex); &
0.243 \tikz\draw[second,fill=second,opacity=0.,fill opacity=0.](0,0)circle(.5ex); &
0.137 \tikz\draw[second,fill=second,opacity=0.,fill opacity=0.](0,0)circle(.5ex); &
17.93 \tikz\draw[second,fill=second,opacity=0.,fill opacity=0.](0,0)circle(.5ex); &
0.701 \tikz\draw[second,fill=second,opacity=0.,fill opacity=0.](0,0)circle(.5ex); &
0.258 \tikz\draw[second,fill=second,opacity=0.,fill opacity=0.](0,0)circle(.5ex); &
0.132 \tikz\draw[second,fill=second,opacity=0.,fill opacity=0.](0,0)circle(.5ex); \\
SplatFields~\cite{mihajlovic2025splatfields}  &
17.81 \tikz\draw[second,fill=second,opacity=0.,fill opacity=0.](0,0)circle(.5ex); &
0.739 \tikz\draw[second,fill=second,opacity=0.,fill opacity=0.](0,0)circle(.5ex); &
0.201 \tikz\draw[second,fill=second,opacity=0.,fill opacity=0.](0,0)circle(.5ex); &
0.119 \tikz\draw[second,fill=second,opacity=0.,fill opacity=0.](0,0)circle(.5ex); &
17.74 \tikz\draw[second,fill=second,opacity=0.,fill opacity=0.](0,0)circle(.5ex); &
0.682 \tikz\draw[second,fill=second,opacity=0.,fill opacity=0.](0,0)circle(.5ex); &
0.207 \tikz\draw[second,fill=second,opacity=0.,fill opacity=0.](0,0)circle(.5ex); &
0.125 \tikz\draw[second,fill=second,opacity=0.,fill opacity=0.](0,0)circle(.5ex); \\
\rowcolor{myrowcolor}
\method &
\textbf{20.37} \tikz\draw[second,fill=second,opacity=0.,fill opacity=0.](0,0)circle(.5ex); &
\textbf{0.797} \tikz\draw[second,fill=second,opacity=0.,fill opacity=0.](0,0)circle(.5ex); &
\textbf{0.143} \tikz\draw[second,fill=second,opacity=0.,fill opacity=0.](0,0)circle(.5ex); &
\textbf{0.084} \tikz\draw[second,fill=second,opacity=0.,fill opacity=0.](0,0)circle(.5ex); &
\textbf{20.62} \tikz\draw[second,fill=second,opacity=0.,fill opacity=0.](0,0)circle(.5ex); &
\textbf{0.762} \tikz\draw[second,fill=second,opacity=0.,fill opacity=0.](0,0)circle(.5ex); &
\textbf{0.179} \tikz\draw[second,fill=second,opacity=0.,fill opacity=0.](0,0)circle(.5ex); &
\textbf{0.091} \tikz\draw[second,fill=second,opacity=0.,fill opacity=0.](0,0)circle(.5ex); \\
\bottomrule
\end{tabular}
\label{tab:tab_real_results}
\end{table*}

\subsection{Results on synthetic scenes}
\textbf{Implementation Details} For synthetic evaluation, we use the \dataset and the CarPatch~\cite{di2023carpatch} dataset as benchmarks to assess 3D reconstruction performance. To simulate a sparse input scenario during training, we sampled 4 images from the training dataset. All images have a resolution of $800\times800$ and are chosen from distinct viewpoints around the vehicles to ensure comprehensive visibility of the object's structure. The $h$ factor of SLERP (Eq.~\ref{eq:slerp}) varies from 0.025 to 0.975 (included) with a step size of 0.025. This results in a total of 156 novel views for a scene with only 4 images.
Ground-truth (GT) camera poses and GT depth maps are employed during the image augmentation phase (Sec. \ref{sec:augmentation}) to enhance rendering accuracy.
To ensure fair comparison, SplatFields is evaluated using GT camera poses, while DNGaussians employs also GT depth maps directly, bypassing the DPT~\cite{ranftl2021vision} computation suggested in the original work. Given that all these methods utilize the 3DGS codebase, we conducted experiments in the synthetic scenario by initializing the point cloud for 3DGS with random values. This approach ensures a fair comparison of their performance under identical initial conditions.

\tit{Main Results} As demonstrated in Table~\ref{tab:tab_synt_results}, \method exhibits significant advantages in the context of novel view synthesis in challenging sparse-view scenarios. The results clearly demonstrate the effectiveness of \method compared to other state-of-the-art methods.
For the \dataset and CarPatch datasets, \method consistently improves all the considered metrics. These results highlight \method's ability to preserve perceptual quality and geometric accuracy, while demonstrating scalability and generalization across diverse scenes.
Notably, 3DGS, while competitive in some cases, falls short in preserving perceptual details (higher LPIPS) and maintaining geometric accuracy (higher AVGE) compared to \method. 


\tit{Additional Analysis}  
Figure \ref{fig:n_images_plot} compares the average error (AVGE) of \method with other approaches under varying numbers of input views. The results demonstrate that our method consistently outperforms the others, particularly in sparse view scenarios with 4 or 8 input images, highlighting its capability to reconstruct 3D scenes effectively even in under-constrained conditions. As the number of input images increases to 12, the AVGE converges with that of SplatField and 3DGS, showcasing its robustness and scalability to denser input settings. In terms of runtime, our approach introduces an extra preprocessing overhead ranging from 25 seconds to 10 minutes, depending on the number of rendered images. For sampling and rendering novel views, it does not affect memory usage, which remains consistent. \method strikes a balance by achieving superior reconstruction quality compared to 3DGS while maintaining a reasonable computation time.

 \begin{figure*}[ht]
     \centering
     \includegraphics[page=4 ,width=\linewidth]{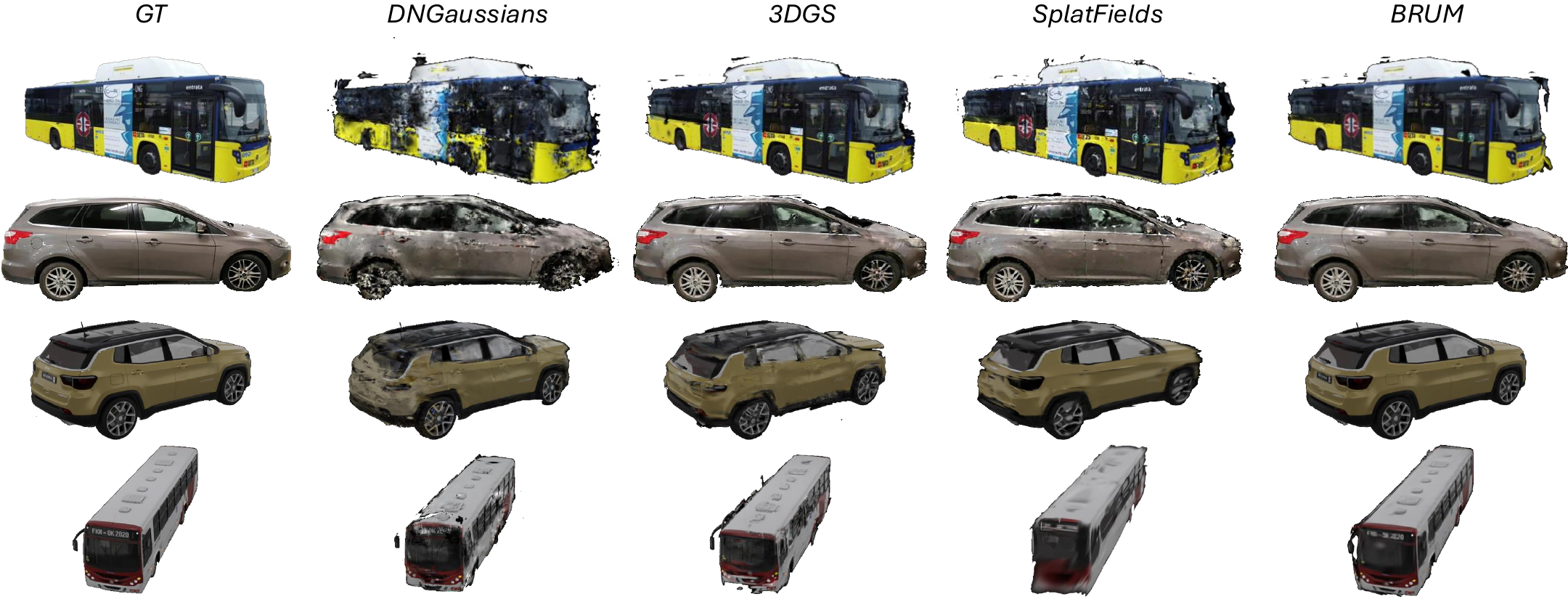}
     \caption{Qualitative comparison of 3D reconstruction methods for DNGaussian, 3DGS, SplatFields, and \method.}
     \label{fig:qualitativi}
 \end{figure*}

\subsection{Results on real-world scenes}
\tit{Implementation details} For the real-world scenario, we evaluate our method on the \dataset and KRONC-dataset \cite{di2024kronc}. Unlike the synthetic case, real-world datasets lack ground-truth camera poses and depth maps. We adopt DUSt3R as a preprocessing step to estimate these parameters, as mentioned in section \ref{sec:DUSt3R}. However, the uncertainties in this estimation process pose challenges for accurate 3D reconstruction. We relax the 4-view constraint and train with 8 views to compensate for inaccuracies in camera pose and depth estimation.

To minimize rendering distortions caused by imprecisions in the depth maps estimated by DUSt3R, the $h$ factor in SLERP (Eq.~\ref{eq:slerp}) for real-world scenes is restricted to a narrower range compared to synthetic scenes, while maintaining a fixed step size of 0.025. Specifically, for the KRONC-dataset $h$ is fixed at 0.1 for all experimentis while for the \dataset is fixed to 0.08. During the preprocessing step (Sec. \ref{sec:DUSt3R} ), we examine the influence of DUSt3R's confidence parameter \(c\). For the KRONC-dataset, the optimal $c$ is found to be 0, allowing all points predicted by DUSt3R to be utilized. In contrast, for the \dataset, the optimal $c$ is determined to be 1.5. In the following sections, we analyze how varying $h$ and $c$ influence reconstruction metrics.
To meet DUSt3R’s requirements, all input images were downsampled to \(512 \times 256\). 
\definecolor{myrowcolor}{HTML}{F5F5F5}
\definecolor{first}{HTML}{dec546}
\definecolor{second}{HTML}{d7d7d7}
\definecolor{third}{HTML}{7e4205}

\begin{table}[t]
\centering
\caption{Quantitative results for different interpolation factors, computed on the Ford scene of the KRONC dataset.}

\begin{tabular}{c|ccc|c}
\toprule
\textbf{$h$} & \textbf{PSNR}~$\uparrow$ & \textbf{SSIM}~$\uparrow$ & \textbf{LPIPS}~$\downarrow$ & \textbf{AVGE}~$\downarrow$\\
\midrule
 0.05 & 21.01 & 0.824 & 0.125 & 0.075\\
\rowcolor{myrowcolor}
\textbf{0.1} & \textbf{21.18} & \textbf{0.827} & \textbf{0.114} & \textbf{0.071}\\
0.3 & 21.13 & 0.827 & 0.120 & 0.073\\
\rowcolor{myrowcolor}

0.5 & 20.95 & 0.823 & 0.119 & 0.074\\
\bottomrule
\end{tabular}
\label{tab:int_factors}
\end{table}

\tit{Main Results} As shown in Table~\ref{tab:tab_real_results}, \method exhibits strong performance in real-world scenarios, though slightly less pronounced than in the synthetic setting (Table~\ref{tab:tab_synt_results}). Real-world datasets present additional challenges, including noise, reflections, and inaccuracies in estimated depth maps and camera poses. Despite these obstacles, \method achieves the highest scores across both datasets. Notably, neither DNGaussians nor SplatFields were originally designed to handle sparse 360° forward-facing real scenes in their formulations. This explains their lower performance highlighting the difficulty of our task. 

\tit{Additional analysis} The results in Table~\ref{tab:int_factors} provide valuable insights into the effect of the maximum interpolation factor $h$ on \method's performance in real-world scenarios. Here, $h$ represents the upper limit for interpolation, while the step size remains fixed at 0.025. Higher values of $h$ indicate moving farther away from the ground truth camera position, resulting in the generation of more synthetic views. For smaller values of $h$, such as 0.05 and 0.1, \method achieves superior preservation of image quality compared to the ground truth. However, as $h$ increases (\textit{e.g.}, 0.3 and 0.5), AVGE experiences a marginal increase. This suggests that larger $h$ values may amplify distortions caused by inaccuracies in depth map and camera pose estimations. 
\definecolor{myrowcolor}{HTML}{F5F5F5}
\definecolor{first}{HTML}{dec546}
\definecolor{second}{HTML}{d7d7d7}
\definecolor{third}{HTML}{7e4205}

\begin{table}[t]
\centering
\caption{Performance of \method evaluated with varying $c$ (Eq.~\ref{eq:point_cloud_filter}) on the bus\_1 scene of the \dataset.}
\begin{tabular}{c|ccc|c}
\toprule
\textbf{$c$} & \textbf{PSNR}~$\uparrow$ & \textbf{SSIM}~$\uparrow$ & \textbf{LPIPS}~$\downarrow$ & \textbf{AVGE}~$\downarrow$\\
\midrule
0   & 17.82 & 0.754 & 0.200 & 0.118 \\
\rowcolor{myrowcolor}
0.5 & 17.81 & 0.755 & 0.199 & 0.118 \\
1   & 17.67 & 0.752 & 0.201 & 0.120 \\
\rowcolor{myrowcolor}
\textbf{1.5} & \textbf{18.41} & \textbf{0.789} & \textbf{0.167} & \textbf{0.103} \\
2   & 18.13 & 0.787 & 0.169 & 0.106 \\
\rowcolor{myrowcolor}
2.5 & 18.18 & 0.785 & 0.170 & 0.106 \\
\bottomrule
\end{tabular}
\label{tab:threshold_metrics}
\end{table}

Table \ref{tab:threshold_metrics} examines the impact of $c$ (Eq.~\ref{eq:point_cloud_filter}) on reconstruction performance using the bus\_1 scene from the \dataset. Higher $c$ values remove more points from the DUSt3R point cloud, leaving larger image portions unprojected. Lower $c$ values (\textit{e.g.}, 0 or 0.5) fail to filter depth estimation uncertainties, while higher values (\textit{e.g.}, 2 or 2.5) overly prune the point cloud, causing information loss. A value of $c = 1.5$ provides the best balance, improving reconstruction accuracy.

Finally, Table~\ref{tab:config_results} evaluates the impact of \method's key components, as outlined in Section~\ref{sec:augmentation}.
The most significant degradation occurs when the XNOR operation from Eq.~\ref{eq:xnor} is removed, and the original foreground mask $\hat{V}$ is used. Additionally, omitting the weighting factor $\tilde{W}$ or reintroducing $\mathcal{L}_{SSIM}$ for generated images in Eq.~\ref{eq:final_loss} results in noticeable performance drops. The complete method achieves the best metrics, validating the importance of each component.
 
\definecolor{myrowcolor}{HTML}{F5F5F5}
\definecolor{first}{HTML}{dec546}
\definecolor{second}{HTML}{d7d7d7}
\definecolor{third}{HTML}{7e4205}


\begin{table}[t]
\centering
\caption{Quantitative results for various configurations. The best-performing corresponds to \method, as described in Sec.~\ref{sec:augmentation}, where the SSIM loss is excluded during training.}
\resizebox{\linewidth}{!}{ 
\begin{tabular}{ccc|ccc|c}
\toprule
\textbf{$\bm{\mathcal{L}_{SSIM}}$} & \textbf{XNOR} & \textbf{$\bm{\hat{W}}$} & \textbf{PSNR}~$\uparrow$ & \textbf{SSIM}~$\uparrow$ & \textbf{LPIPS}~$\downarrow$ & \textbf{AVGE}~$\downarrow$\\
\midrule
\cmark & \xmark & \cmark & 20.07 & 0.796 & 0.148 & 0.087\\
\rowcolor{myrowcolor}
\cmark & \cmark & \xmark & 20.46 & 0.798 & 0.131 & 0.081\\
\cmark & \cmark & \cmark & 20.95 & 0.808 & 0.128 & 0.077\\
\rowcolor{myrowcolor}
\xmark & \cmark & \cmark & \textbf{21.18} & \textbf{0.827} & \textbf{0.116} & \textbf{0.072}\\
\bottomrule
\end{tabular}
\label{tab:config_results}}
\end{table}

\subsection{Qualitative results}
In Fig.~\ref{fig:qualitativi}, we compare qualitative results on the \dataset, KRONC, and CarPatch datasets. The proposed method accurately reconstructs vehicles in both synthetic and real domains, preserving fine details and producing outputs closer to ground truth data. 

\section{Conclusion}
This paper addresses the challenge of 3D vehicle reconstruction from sparse views. By leveraging depth maps and the DUSt3R architecture, \method significantly enhances the robustness of Gaussian Splatting in scenarios with limited input data. We introduce a novel dataset combining synthetic and real-world public transport vehicles, achieving high-quality reconstructions across challenging scenes. The scalability and adaptability of our approach make it well-suited for practical deployment in resource-constrained environments.

\section*{ACKNOWLEDGMENT}
This research was partially funded by the International Foundation Big Data and Artificial Intelligence for Human Development (IFAB). The authors also acknowledge Società Emiliana Trasporti Autofiloviari (SETA) S.p.A. based in Modena for allowing to capture the real public transportation scenes of the {\dataset} featured in this work.

\thispagestyle{empty}
\pagestyle{empty}

\bibliographystyle{IEEEtran}
\bibliography{main}

\end{document}